\documentclass[conference]{IEEEtran}
\IEEEoverridecommandlockouts
\usepackage{cite}
\usepackage{amsmath,amssymb,amsfonts}
\usepackage{algorithmic}
\usepackage{graphicx}
\usepackage{textcomp}
\usepackage{url}
\usepackage{xcolor}
\usepackage{dblfloatfix}
\def\BibTeX{{\rm B\kern-.05em{\sc i\kern-.025em b}\kern-.08em
    T\kern-.1667em\lower.7ex\hbox{E}\kern-.125emX}}
\begin{document}

\title{Characteristic Energy Behavior Profiling of Non-Residential Buildings\\

\author{\IEEEauthorblockN{1\textsuperscript{st} Haley Dozier}
\IEEEauthorblockA{\textit{Information Technology Laboratory} \\
\textit{U.S. Army Engineer Research and Development Center}\\
Vicksburg, M.S. U.S.A \\
Haley.R.Dozier@usace.army.mil}
% \and
% \IEEEauthorblockN{2\textsuperscript{nd} Indu Shukla}
% \IEEEauthorblockA{\textit{Information Technology Laboratory} \\
% \textit{U.S. Army Engineer Research and Development Center}\\
% Vicksburg, M.S. U.S.A \\
% Indu.Shukla@usace.army.mil}

\and
\IEEEauthorblockN{2\textsuperscript{nd} Althea Henslee}
\IEEEauthorblockA{\textit{Information Technology Laboratory} \\
\textit{U.S. Army Engineer Research and Development Center}\\
Vicksburg, M.S. U.S.A \\
Althea.C.Henslee@usace.army.mil}
% \and
% \IEEEauthorblockN{3\textsuperscript{rd} Alexandre Ligo}
% \IEEEauthorblockA{\textit{Environmental Laboratory Laboratory} \\
% \textit{U.S. Army Engineer Research and Development Center}\\
% Concord, M.A. U.S.A \\
% Alexandre.Ligo@usace.army.mil}
% \and
% \IEEEauthorblockN{4\textsuperscript{th} Steffenie Fries}
% \IEEEauthorblockA{\textit{Environmental Laboratory} \\
% \hspace{1cm}\textit{U.S. Army Engineer Research and Development Center}\hspace{1cm}\\
% ??? U.S.A \\
% ???@usace.army.mil}

% \and
% \IEEEauthorblockN{5\textsuperscript{th} Igor Linkov}
% \IEEEauthorblockA{\textit{Environmental Laboratory} \\
% \textit{U.S. Army Engineer Research and Development Center}\\
% Concord, M.A. U.S.A \\
% Igor.Linkov@usace.army.mil}
}

}

\maketitle

\begin{abstract}
Due to the threat of changing climate and extreme weather events, the infrastructure of the United States Army installations is at risk. More than ever, climate resilience measures are needed to protect facility assets that support critical missions and help generate readiness. As most of the Army installations within the continental United States rely on commercial energy and water sources, resilience to the vulnerabilities within independent energy resources (electricity grids, natural gas pipelines, etc) along with a baseline understanding of energy usage within installations must be determined. This paper will propose a data-driven behavioral model to determine behavior profiles of energy usage on installations. These profiles will be used 1) to create a baseline assessment of the impact of unexpected disruptions on energy systems and 2) to benchmark future resiliency measures. In this methodology, individual building "behavior" will be represented with models that can accurately analyze, predict, and cluster multimodal data collected from energy usage of non-residential buildings. Due to the nature of Army installation energy usage data, similarly structured open access data will be used to illustrate this methodology.  
\end{abstract}


\section{Introduction}

% Climate change is an emerging issue withing the United States Department of Defense (DoD) that presents potentially major national security threats and implications. 

Spending more than $1$ billion per year on facility energy and water, the U.S. Army is the greatest consumer of installation energy within the Department of Defense \cite{resilient_installations_article}. Although this cost seems monumental to some, uninterrupted access to both energy and water is imperative to the Army's requirement to complete critical missions and is essential to readiness. Vulnerabilities in the commercial energy resources that many installations rely on can jeopardize an installations ability to operate and accomplish its mission. Therefore, an installations energy resilience, or ability to both detect and recover from outages, is crucial to the Army's ability to deploy, fight, and win.

% In a speech to the Leaders Summit on Climate, Secretary of Defense Lloyd J. Austin III stated "Today, no nation can find lasting security without addressing the climate crisis. We face all kinds of threats in our line of work, but few of them truly deserve to be called existential. The climate crisis does" \cite{Sec_Def_speech}.  

The development of modernized installations with systems that are resilient to disruption requires research and development in multiple areas that support varying levels of strategic outcomes
\cite{Army_installation_strat}. In this paper, we focus on a small piece of the resiliency puzzle - the determination and clustering of energy behavioral profiles. Once created, these profiles can be used to not only detect deviation from normal but also to provide initial determinations of the current state or baseline of resiliency for energy resiliency. This baseline can then be used to assess the effectiveness of future resiliency measures. Additionally, unsupervised dimension reduction and clustering on these profiles can determine groups of buildings with similar energy consumption. The grouping of these similar entities can be used to predict consumption and resiliency behavior at locations in which little data is currently collected. Clustering can also help identify areas or buildings that may be at higher risk from extreme weather events such as flooding or storms due to their proximity to water sources or other climatic factors. Knowing this information allows military personnel and administrators to better prepare for potential disasters before they happen, making sure that necessary supplies and manpower are available.

\begin{figure*}[!b]
    \centering
    \includegraphics[width = .95\textwidth]{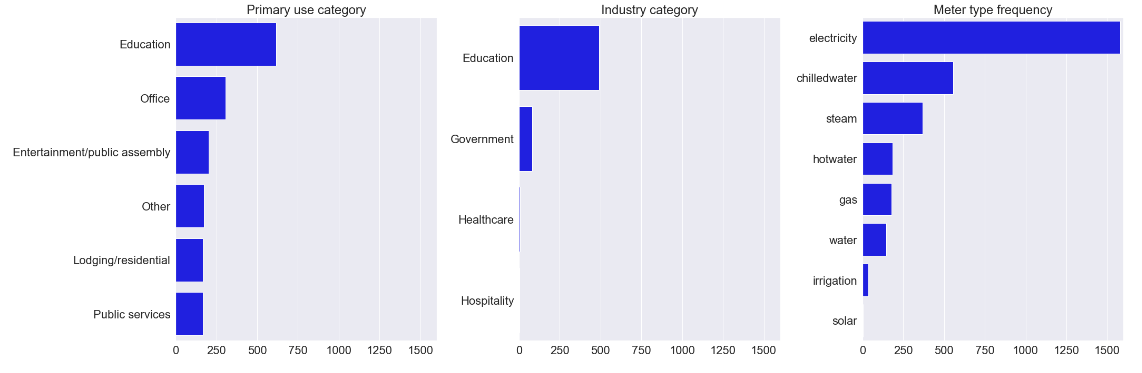}
    \caption{Top primary usage, industry usage, and meter types in the Building Data Genome 2 data set.}
    \label{fig:Industry_and_use}
\end{figure*}

This paper will proceed as follows. Section \ref{sec:background} describes the Army Central Metering Program. Section \ref{sec:Materials_and_Methods} will overview the data used for this analysis as well as the methodology. Section \ref{sec:Results_and_Discussion} will describe the results obtained from using the methodology outlined in the previous section. And finally, section \ref{sec:Conclusions_and_FutureWork} will detail the conclusions derived from the results as well as outline potential future directions for this effort.

It should be noted that due to the sensitive nature and availability of Army installation energy usage data, we focus on the creation and clustering of the energy consumption profiles using open access data that is similarly structured to energy usage data that can be obtained from installations.

%this should be in the next section but i am putting it here to help with formatting

\section{Background: Army Central Metering Program} \label{sec:background}

Through the Army Central Metering Program (ACMP), select Army facilities have been collecting metering data since 2005. In 2011, per executive order 028-12, all Army facilities will work with the designated center of expertise (U.S. Army Corps of Engineers - Engineering and Support Center in Huntsville, AL) to incorporate energy monitoring systems and per Army Directive 2014-10 (Advanced Metering of Utilities), the metering systems must be connected to the Army Meter Data Management system. \cite{Army_Central_Metering_Program}
% ACMP requires all installed meters that are part of the reporting infrastructure meet or exceed all specifications stated by the contract regardless of the source or ownership of the meters. If an existing meter does not meet the minimum specifications in all areas, then it will be replaced, or if replacement is impractical due to ownership or other contractual issues, an additional meter will be installed on the load side of any existing meter. Regardless of the funding source or contract vehicle used to procure/install, all new meters shall meet all specifications required by the metering program to include compatibility with the planned Army meter data network and full, unconditional access to the meter data for which the Army has responsibility and accountability irrespective of metering equipment/network ownership.

% Center of Expertise = U.S. Army Corps of Engineers - Engineering and Support Center, Huntsville \cite{Army_Central_Metering_Program}

\subsection{Army Meter Data Management System}

In support of the ACMP, The U.S. Army Corps of Engineers (USACE) provides maintenance, accreditation and operation of the Army Meter Data Management System (MDMS). The Army Meter Data Management System \cite{MDMS} was developed to monitor the United States Army's global consumption of energy and water. The MDMS Enterprise uses a cental database to automate meter data collection on a secure network, and generates energy reports that are accessible via the MDMS Enterprise Portal. By using the MDMS Enterprise, Army installations can track utility commodity consumption at the facility level. As of 2021, there are over 17,000 meters (electricity, gas, steam, water, etc.) from different installations reporting to the MDMS with plans to integrate even more \cite{Army_Central_Metering_Program, MDMS_news1}. 

In addition to tracking energy use, when the data from the MDMS is collected and reported it can often  help installations find and fix system problems, track down the source of power outages, or detect cyber attack when unexpected spikes occur. But, a significant challenge to the MDMS enterprise is uninterrupted collection of the necessary energy metering data to inform installations of these energy issues. Energy meter data at some installation is often collected by installation employees or energy managers. This system of data collection often leads to error in collection and large, unexplainable gaps in the data.

% Smart met

% For example, during the global pandemic due to the SARS-CoV-2 virus, many installations had to reduce workplace operations.  

% This data is often collected by energy managers or smart meters.

% In addition to tracking energy use, the data from the MDMS can often provide explanations or help installations find and fix system problems when unexpected spikes occur. 

% Trudell said the system has already helped customers track down the source of power outages at two facilities and identified the cause of a spike in water use at another.

% WRITE ABOUT HOW THEY ARE GETTING MORE DATA AND MAINTAINING THE INSTALLATION LEVEL MANAGEMENT CENTER BUT A LOT OF HISTORICAL DATA IS MISSING AND METERS ARE BAD 

% Link: \url{https://apps.dtic.mil/sti/citations/ADA565695}

\subsection{Army Energy and Water Reporting System)}

The Army Energy and Water Reporting System (AEWRS) contains Army installation energy consumption data. Each installation is responsible for inputting accurate data each month. Various government offices for energy conservation evaluation and other decision makers can then access this information \cite{AEWRS1}. 
% Authorized users have access to more than 50 AEWRS/Energy Manager reports. Reports can be shown for a specific installation, region, Land-holding command, Army Component, or Army wide. Authorized AEWRS users have access to more than 50 Energy Manager reports that can be shown for a specific installation, region, Land-holding command, Army Component, or Army wide \cite{AEWRS1}.

\section{Materials and Methods} \label{sec:Materials_and_Methods}
\subsection{Description of Data}

As previously stated, due to the sensitive nature of the data collected from military installations, for the results presented in this paper we use an open-access dataset with similar characteristics to those obtained from installation specific data sets such as the metering data on the Army Meter Data Management System (MDMS) or Army Energy and Water Reporting System (AEWRS).

\subsubsection{The Building Data Genome 2 Data-Set}

\begin{figure}[!ht]
    \centering
    \includegraphics[width = .45\textwidth]{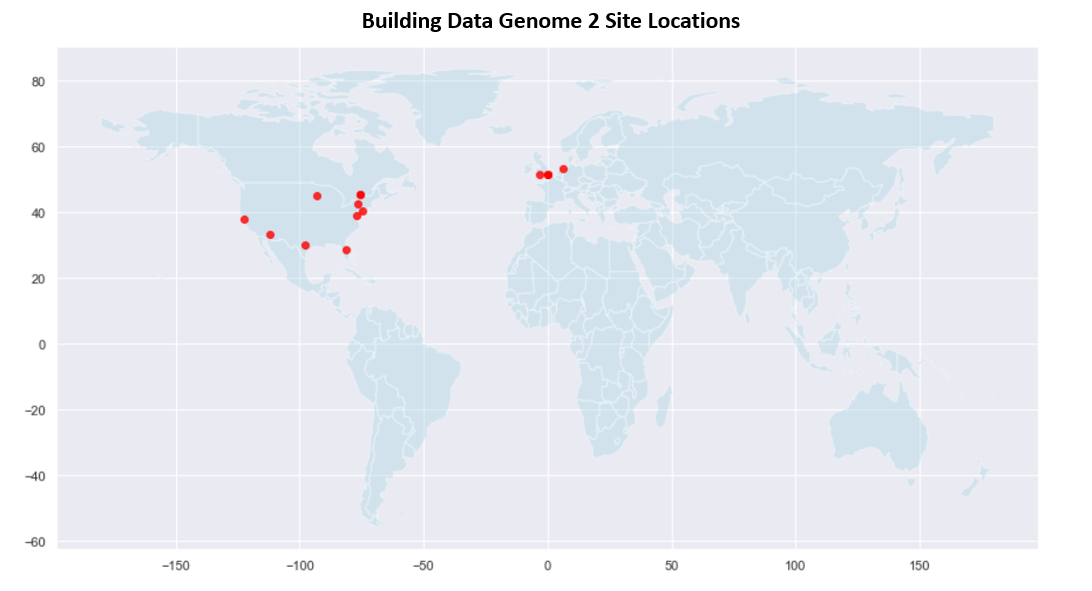}
    \caption{Locations of the building sites included in Building Data Genome 2 Data-Set. Each site is designated by a red dot. }
    \label{fig:Genome_Map}
\end{figure}

To effectively apply an eigendecomposition of a data set, a large repository of temporal data is needed. The Building Data Genome 2 (BDG2) Data-Set \cite{Genome_dataset} is an open-access dataset containing energy meters from 1,636 buildings across 19 sites located in the United States, Canada, United Kingdom, and Ireland (see Figure \ref{fig:Genome_Map}). This dataset includes a variety of metadata for the different buildings. This metadata can include primary usage, timezones, building size (square feet and square meters), year built, and weather information. Figure \ref{fig:Industry_and_use} displays the distribution of data within select metadata categories for the dataset. 

% industry including government, education, healthcare, and hospitality with 16 types of usage (office, entertainment, lodging, etc). 

Similar to Army MDMS and AEWRS data, the BDG2 dataset contains whole-building data for from electricity,  steam,  gas, and water meters from select sites. Unlike MDMS and AEWRS, the BDG2 dataset also contains chilled water, hot water, irrigation, and solar meters. Additionally, the BDG2 dataset was only collected over a two year time period, is reported in hourly increments instead of every 15 minutes, and includes more complete and specific metadata describing the usage of each building.

\subsection{Methodology}

The methodology of this work centers around using eigenbehavior analysis along with unsupervised machine learning techniques to understand the characteristic behaviors of the energy usage of a building. 

\subsubsection{Eigenbehavior Analysis}

The goal of eigenbehavior analysis is to identify typical behavioral profiles within a dataset. Eagle and Pentland \cite{Eigenbehavior_people} first suggested Eigenbehavior analysis for discovering behavioral patterns in the location data of 100 people monitored by smartphone communications, and over the years this approach has been tested on other areas of application such as typical operating strategies across a large set of water reservoirs \cite{Eigenbehaviour_water2}, behavior of an articulated body \cite{Eigenbehaviour_robots}, and the cognitive ability in older adults \cite{Eigenbehavior_cognition}. 

\begin{figure*}
    \centering
    \includegraphics[width = .45\textwidth]{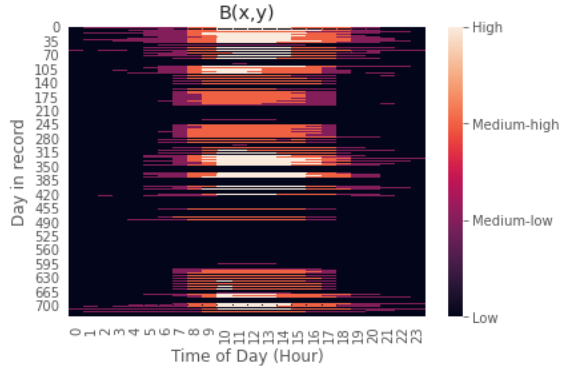}
    \includegraphics[width = .48\textwidth]{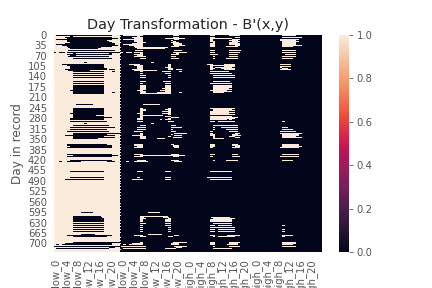}
    \centering
    \caption{Tranformation from $B(x,y)$ to $B'(x,y)$ for a single building. The left plot represents the energy consumption over the course of 2 years for the different levels of categorical consumption. This data is then transformed into the binary matrix $B'(x,y)$ on the right. Color intensity is proportional to the magnitude of each entry. }
    \label{fig:B_x_y_and_transformation}
\end{figure*}

To identify the characteristic energy usage of a set of buildings, first categories of energy usage must be developed. For this work, we will categorize the energy usage into $n=4$ categories of energy behavior: low, medium-low, medium-high, and high. Then we follow the framework of Eagle and Pentland by characterizing each entity, or building, by a two-dimensional $D \times 24 $ array of usage information, $B(x,y)$ where $D$ is the number of days in the dataset (see Figure \ref{fig:B_x_y_and_transformation}). This array contains the predefined labels corresponding to a behavior over the $24$ hours in the $D$ number of days on record. Then, to perform the analysis, $B(x,y)$ is transformed to a binary array, $B'(x,y)$, representing the behavior of the energy usage over a day (see Figure \ref{fig:B_x_y_and_transformation}). This transformed matrix is of size $D$ by $96$ where $96$ is equal to the number of hours in a day times the number of possible labels ($n=4$). Each row, $\Gamma_i$, of the transformed binary matrix, $B'(x,y)$ , represents the categorized energy usage over day $i\epsilon[1,2,...,D]$ and contains $96$ entries.

% \begin{figure}
%     \centering
    
%     \caption{Caption}
%     \label{fig:Binary_array}
% \end{figure}

% , of size days in the dataset by hours in a day $(D \times 24)$

From here, an eigendecomposition of $B'(x,y)$ is performed to determine the eigenvalues and eigenvectors of the array. That is, given a set of each day's energy usage behaviors $\Gamma_1. \Gamma_2,..., \Gamma_D$ the average behavior of the individual is calculated by
\begin{equation} \label{eqn:average_behavior}
    \mu = \frac{1}{D}\sum^{D}_{i=1}\Gamma_i 
\end{equation}
and the deviation in the behavior of the building in a day is calculated using 
\begin{equation} \label{eqn:deviation}
    \Phi_i = \Gamma_i - \mu. 
\end{equation}
Using equation \ref{eqn:average_behavior}, the covariance matrix is computed where each entry satisfies
\begin{equation}
    % cov_{x,y}=\frac{\sum_{i=1}^{N}(x_{i}-\bar{x})(y_{i}-\bar{y})}{N-1}
    C = \frac{1}{H}\sum_{i=1}^{H}\Phi_i\Phi_{i}^T
\end{equation}
where $H$ is the number of hours in a day times the number of possible labels ($24\times n$) \cite{Eigenbehavior_people}. The eigenvalues, $\lambda$ are the roots of the characteristic equation $\det(C-\lambda I) = 0$ where $I$ is the identity matrix. Then the corresponding eigenvectors, $\textbf{v}$, are found by solving the equation 
\begin{equation}
    (C-\lambda I)\textbf{v}=0.
\end{equation}

Due to the amount of structure in the daily operation of an installation or non-residential building, daily energy consumption of buildings should be clustered. This allows for the creation of a low-dimensional "behavior space" that can be defined by a subset of vectors that best characterize the distribution of energy consumption and can be referred to as the primary eigenvector, or "eigenbehavior". These are found by ranking the eigenbehaviors by the total amount of variance it accounts for in the data (which is equivalent to the scale of the associated eigenvalue). The primary eigenbehavior of the building is considered to be the eigenbehavior corresponding to the eigenvalue with the highest value \cite{Eigenbehaviour_water1}. This process can be repeated across all buildings with electricity metering in the BDG2 dataset. 

\subsubsection{Clustering for Routine Similarities}

To assess similarities in the main daily energy consumption routines the principle behaviors of the buildings can be clustered based on the similarity of their first eigenbehavior. The clustering of these behaviors will determine the types of daily energy consumption routines in the dataset. 

Due to large dimensionality of the collection of primary eigenbehaviors for each building, to cluster the behaviors into consumption profiles, first a dimension reduction method must be deployed. The method used for this process was the non-linear dimension reduction method Uniform Manifold Approximation and Projection (UMAP) \cite{umap-software} due to this algorithms success in preserving both global and local structure within a manifold. UMAP models high-dimension manifolds with a fuzzy topological structure and then embeds the data by searching for a low dimensional projection that represents the closest equivalent fuzzy topological structure \cite{UMAP}. Once the lower-dimensional representation is obtained, traditional K-means clustering \cite{kmeans} can be implemented to separate the data.

\section{Results and Discussion} \label{sec:Results_and_Discussion}

For select buildings, the BDG2 dataset provides hourly metering data from electricity,  steam,  gas, water, chilled water, hot water, irrigation, and solar meters. Since electricity meters are the most common in the dataset (see Figure \ref{fig:Industry_and_use}), this work will primarily focus on electricity consumption but future work will expand this to include gas, water, and solar. Additionally, it should be noted that before performing the eigendecomposition of the data, the data is scaled by determining the energy use per square foot of the building, then using the Min Max Scaler in the sci-kit-learn library \cite{scikit-learn} to scale these values to between 0 and 1. 

\subsection{Eigenbehavior Analysis}
\begin{figure}
    \centering
    \includegraphics[width = .24\textwidth]{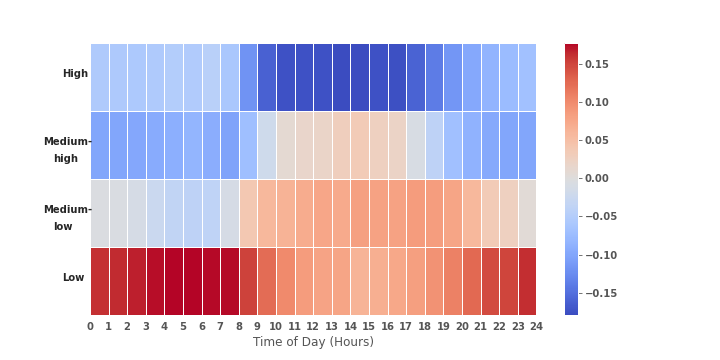}
    \includegraphics[width = .24\textwidth]{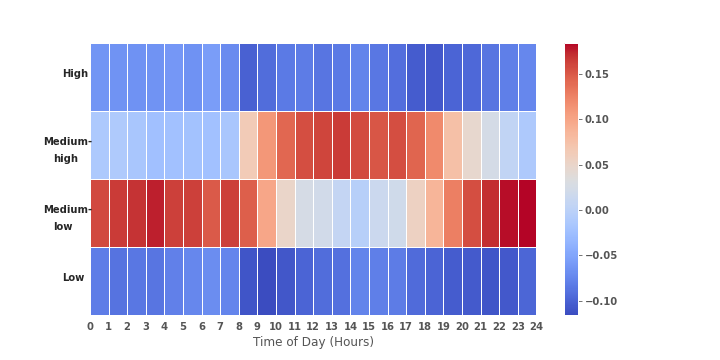}
    \includegraphics[width = .24\textwidth]{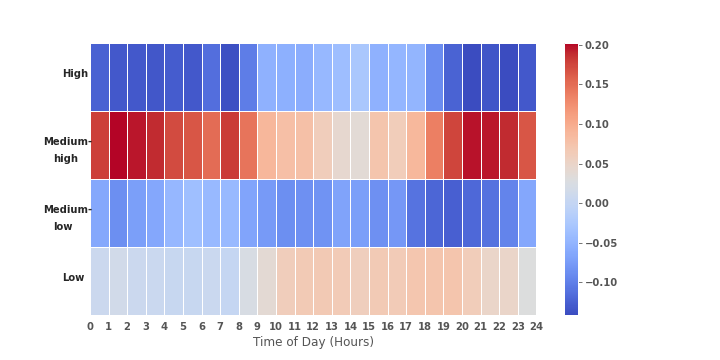}
    \caption{Top three eigenbehaviors for a sample building (left to right). Each eigenbehavior is reshaped to an $4 \times 24$ array so that the behavior can be comprehensively viewed over the course of 24 hours. Color intensity is proportional to the corresponding eigenbehavior entry's magnitude. }
    \label{fig:top_three_behaviors}
\end{figure}
As previously stated, before constructing the eigenbehaviors for a building, first the categories of electricity consumption must be developed. For simplicity, a quantile approach is used relative to each building. Energy consumption in the lowest quarter of data is categorized to be "low" usage, the second quarter of data is categorized to be "medium-low" usage, the third quarter is "medium-high" usage, and the final quarter is "high" usage. Once these categories are defined, the categorical array $B(x,y)$ can be created and then transformed to the binary array $B'(x.y)$ as depicted in Figure \ref{fig:B_x_y_and_transformation}. Each consumption level has $24$ entries representing the hour of each day. Once the binary array is formulated, PCA can be performed to obtain the eigenbehaviors of the energy consumption in a specified building. 

\begin{figure}[!ht]
    \centering
    \includegraphics[width = .4\textwidth]{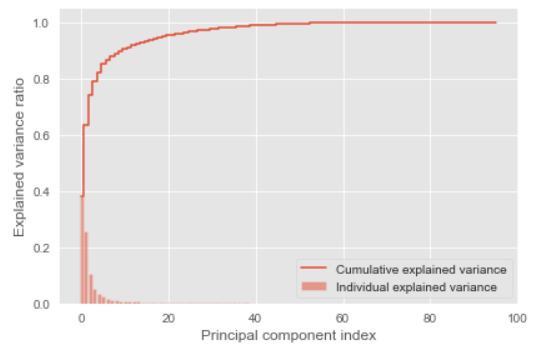}
    \caption{Explained variance computed after using Principle Component Analysis on the transformed binary matrix for the energy consumption of a single building. }
    \label{fig:explained_var}
\end{figure}

The first three eigenbehaviors for a sample building is demonstrated in Figure \ref{fig:top_three_behaviors} and Figure \ref{fig:explained_var} depicts the variance explained by the ranked eigenbehaviors obtained from PCA. As seen in Figure \ref{fig:explained_var}, the first three eigenbehaviors explains approximately $74\%$ of the variance in the electricity consumption data for the sample building. 

\begin{figure}
    \centering
    \includegraphics[width = .5\textwidth]{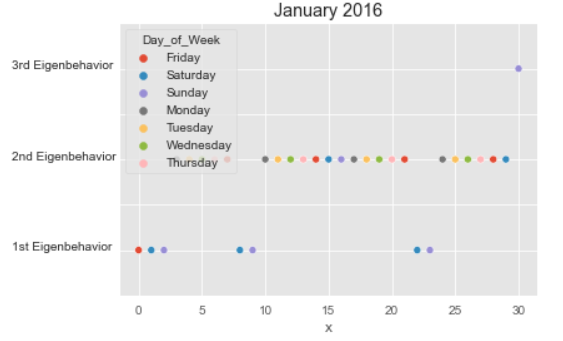}
    \caption{One month (January 2016) of categorized eigenbehaviors.}
    \label{fig:Closest_eigenbehavior_january}
\end{figure}

By calculating the minumum distance between each of the entries of the transformed dataset from PCA and the top three eigenbehaviors, a more comprehensive understanding of the behavior of a single building can be gained. For example, Figure \ref{fig:Closest_eigenbehavior_january} demonstrates that weekdays tend to correspond with the second eigenbehavior while weekends tend to correspond to the first. Repeating this analysis over the entire timeline in the data can give an understanding of how a specific building is being used over the course of the year.

% Before clustering, the primary energy usage eigenbehavior must be collected from each building. 

% \begin{figure}
%     \centering
%     % \includegraphics{Figures/Eigenbehaviors_for_Sampled_buildings.PNG}
%     \includegraphics[width = .55\textwidth]{Figures/Eigenbehaviors_sample.png}
%     \caption{Caption}
%     \label{fig:Eig_buildings_sample}
% \end{figure}

% \textbf{COPIED}
% days are not distributed randomly
% though this large space. Rather, they are clustered, allowing the individual to be described by a relatively low dimensional ‘behavior space’. This space is defined by a subset of vectors of dimension H that can best characterize the distribution of behaviors and are referred to as the primary eigenbehaviors. The top three eigenbehaviors that characterize the individual shown in Fig. 1 are plotted in Fig. 2. The first eigenbehavior corresponds to either a normal day or a day spent traveling (depending on whether the associated weight is positive or negative). The second
% eigenbehavior has a corresponding weight that is positive on weekends and negative on weekdays, analogous to the characteristic behavior of sleeping in and spending that night out in a location besides home or work. The third eigenbehavior is emphasized when the subject is in locations with poor phone reception.
% \textbf{END COPY}

\subsection{Clustering for Routine Similarities}

\begin{figure}
    \centering
    \includegraphics[width = .4\textwidth]{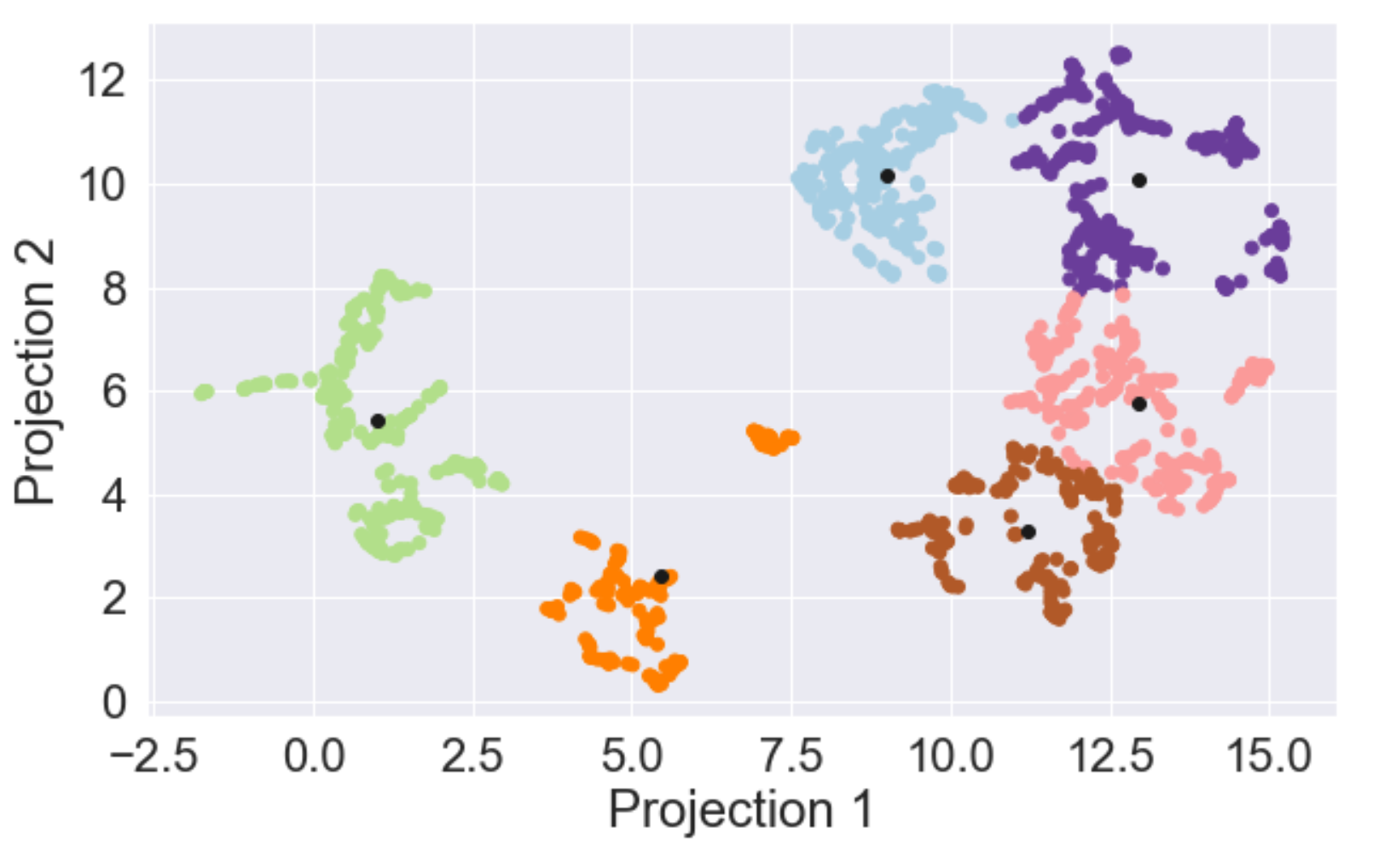}
    \caption{UMAP projection of the primary eigenbehaviors obtained from electricity consumption at each building in the BDG2 dataset. Clusters are determined using K-means clustering and the centroids of each cluster are depicted by black dots.}
    \label{fig:UMAP}
\end{figure}

\begin{figure*}[!ht] %wide figure
    \centering
    \includegraphics[width=.75\textwidth]{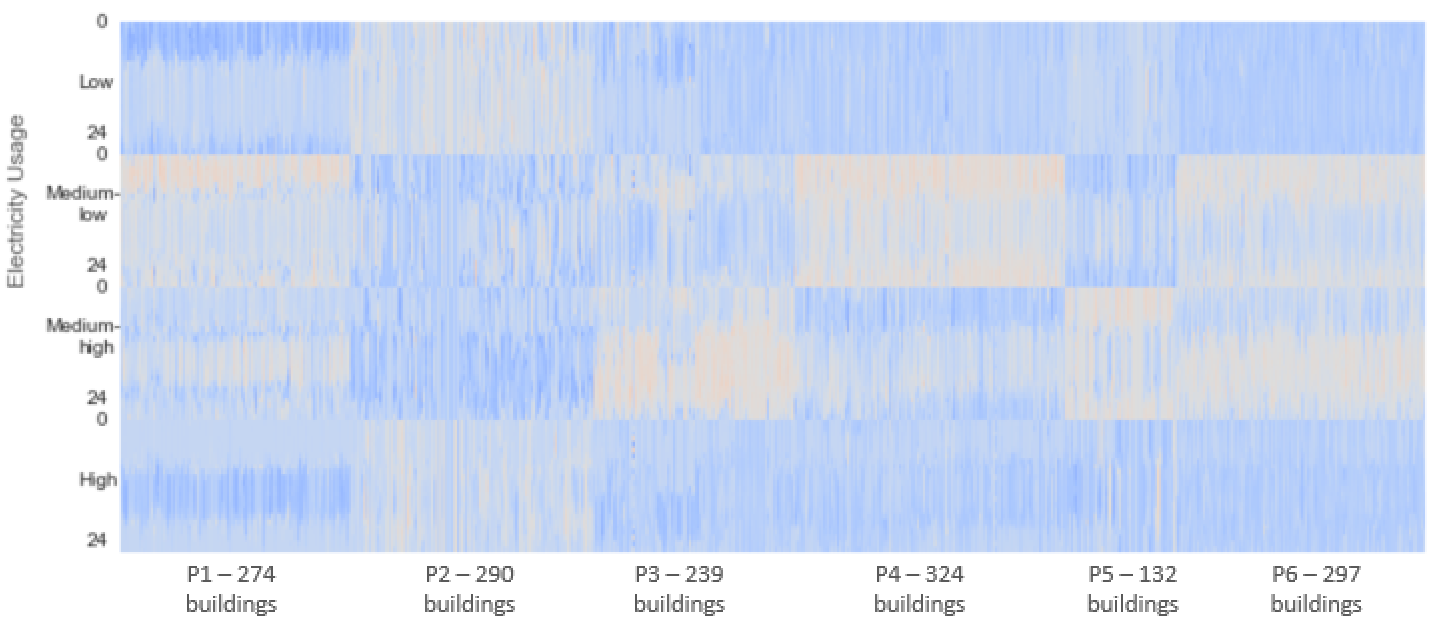}
    \caption{Behavioral profiles (P1, P2, P3, P4, P5, and P6) of the building's electricity usage showing typical behaviors. }
    \label{fig:grouped_behavior}
\end{figure*}

Once the primary eigenbehaviors are computed for each building that contains electricity consumption data in the BDG2 dataset, UMAP can be implemented to obtain a low-dimensional (2-D) representation of the data and K-means clustering can separate similar energy consumption profiles. For this work, 6 behavior profiles were identified: P1, P2, P3, P4, P5, and P6 (see Figures \ref{fig:UMAP} and \ref{fig:grouped_behavior}). 

Among the 6 behavioral profile clusters, typical behaviors which differ in eigenbehavior can be identified. For example, the first profile P1 illustrates electricity usage that varies from medium-low to medium-high usage throughout the day. That is, the highest entries for the behavior represented by P1 move from medium-low to medium-high during the traditional work day, then back to medium-low (similar to the second eigenbehavior in Figure \ref{fig:top_three_behaviors}). This profile is most likely due to a combination of factors, primarily weather, climate, and the standard work schedule observed in the countries represented in the BDG2 dataset. For this first behavioral profile, it is likely that the spike in usage around 9am to 3pm is due to an increase in foot traffic in a building or may correspond to hours of operation.

\begin{figure*}[!ht]
    \centering
    \includegraphics[width = .95\textwidth]{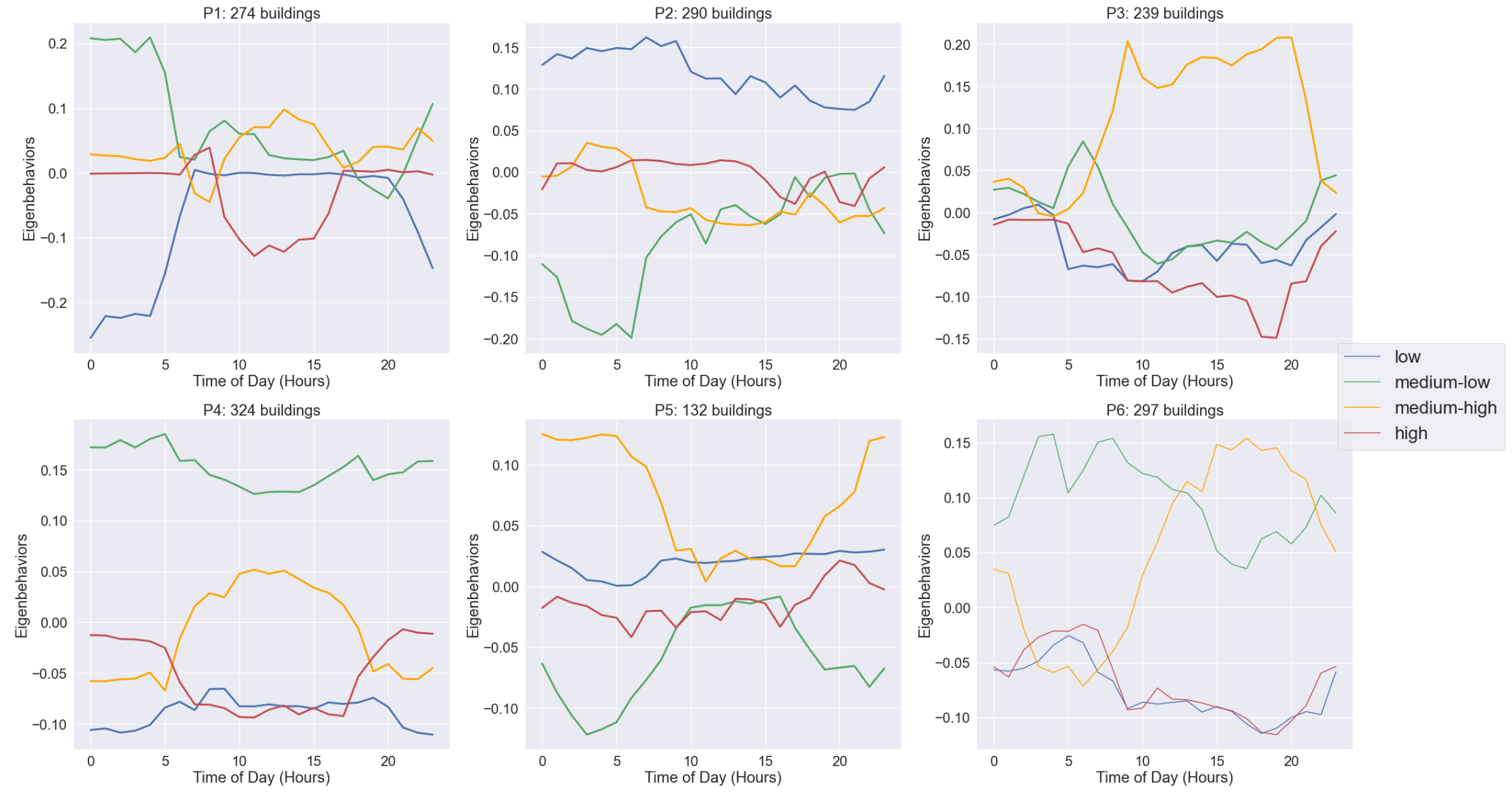}
    \caption{Alternative visualization of the electricity profiles wherein each profile's median eigenbehavior's weights are represented throughout the day. }
    \label{fig:centroid_behaviors}
\end{figure*}

An alternative visualization of the electricity profiles (eigenbehaviors) can be seen in Figure \ref{fig:centroid_behaviors} wherein each profile's median eigenbehavior's weights are represented throughout the day are visualized in a line graph. 

%Haley - you probably need more here

% \subsection{Aligned Uniform Manifold Approximation and Projection}

\section{Conclusions, Future Work, and Recommendations} \label{sec:Conclusions_and_FutureWork}

One advantage to Eigenbehavior analysis is that this methodology gives an understanding of the data that is hard to achieve using alternative unsupervised machine learning alone or other traditional methods. Methods such as Aligned UMAP for time varying data or TimeCluster \cite{TimeCluster} provide a significant amount of promise for visualizing the change in energy consumption over time, but lack clear explainability factors - especially when the clusters do not conform to known labels for the dataset. 

Additionally, eigenbehavior models for a single building can be useful in the case of cyber attack detection or predictive maintenance. Figure \ref{fig:Closest_eigenbehavior_january} demonstrates that there is structure in the routine of electricity consumption in data. With additional information such as planned building usage and weather data, deviation from the established normal behavior for that day can be detected and investigated.  

A surprising result from this methodology is that buildings with in different space usage categories (see Figure \ref{fig:Industry_and_use}) and climates maintain similar consumption routines or energy behavior profiles. In fact, there does not seem to be a strong correlation between industry, space usage category, building size, or location when observing the primary behavioral profile of the building.

\subsection{Future Work}

Overall, the methodology proposed in this paper is suitable for identifying primary behavioral profiles for building, but additional information is required to determine a resiliency baseline for buildings. Data containing electricity outages due to acts of nature, planned maintenance, or attack are needed to fully observe how the eigenbehaviors deviate from what is expected. Although this data can be found in the MDMS and AEWRS datasets, the outage metadata often does not correlate exactly to the energy metering report. Therefore, 

Additionally, the clustering for routine similarity methodology outlined for this work focuses on the primary eigenbehaviors for building energy consumption even though these behaviors may not explain a significant portion of the data. Future work will include additional eigenbehaviors for each building in the assessment. 

Another avenue of future interest is the coupling of additional meters such as gas, solar, and water. This coupling could potentially provide further insights into categorizing energy behavioral profiles and the establishment of a normal operational baseline for consumption.

% \subsubsection{Resilience Baseline}

% FUTURE WORK - ASSESSMENT OF RECOVERY FROM FAILURE - PRETEND WE DID THIS FOR MDMS

% \subsection{Recommendations}

% \section*{Acknowledgment}

% The preferred spelling of the word ``acknowledgment'' in America is without 
% an ``e'' after the ``g''. Avoid the stilted expression ``one of us (R. B. 
% G.) thanks $\ldots$''. Instead, try ``R. B. G. thanks$\ldots$''. Put sponsor 
% acknowledgments in the unnumbered footnote on the first page.

% \section*{References}

\nocite{}


\begin{thebibliography}{99}

\bibitem{Genome_dataset}
Miller, C., Kathirgamanathan, A., Picchetti, B., Arjunan, P., Park, J. Y., Nagy, Z., Raftery, P., Hobson, B. W., Shi, Z., \& Meggers, F. (2020).  
The Building Data Genome Project 2, energy meter data from the ASHRAE Great Energy Predictor III competition.  
\textit{Scientific Data}, 7, 368. Nature Publishing Group.

\bibitem{TimeCluster}
MAli, M., Jones, M. W., Xie, X., et al. (2019).  
TimeCluster: dimension reduction applied to temporal data for visual analytics.  
\textit{The Visual Computer}, 35, 1013–1026.

\bibitem{MDMS}
CALIBRE SYSTEMS INC ALEXANDRIA VA. (2011).  
The Army Meter Data Management System (MDMS): A Case Study For Army MDMS Pilot.  
\textit{Technical Report}. Available: \url{https://apps.dtic.mil/sti/citations/ADA565695}.

\bibitem{AEWRS1}
Army Energy and Water Management Program.  
Army Energy and Water Management Program: Reporting : AEWRS.  
Available: \url{https://army-energy.army.mil/reporting/aewrs.asp}.

\bibitem{Army_Central_Metering_Program}
U.S. Army Corps of Engineers. (2021, August).  
Army Central Metering Program.  
Available: \url{https://www.hnc.usace.army.mil/Media/Fact-Sheets/Fact-Sheet-Article-View/Article/482106/energy-division-army-central-metering-program/}.

\bibitem{Eigenbehaviour_robots}
Jiang, X., \& Motai, Y. (2005).  
Learning by observation of robotic tasks using on-line PCA-based Eigen behavior.  
In \textit{2005 International Symposium on Computational Intelligence in Robotics and Automation} (pp. 391–396). IEEE. doi:10.1109/CIRA.2005.1554308.

\bibitem{Eigenbehavior_people}
Eagle, N., \& Pentland, A. S. (2009).  
Eigenbehaviors: Identifying structure in routine.  
\textit{Behavioral Ecology and Sociobiology}, 63(7), 1057–1066.

\bibitem{Eigenbehaviour_water2}
Giuliani, M., \& Herman, J. D. (2018).  
Modeling the behavior of water reservoir operators via eigenbehavior analysis.  
\textit{Advances in Water Resources}, 122, 228–237.  
doi:10.1016/j.advwatres.2018.10.021.

\bibitem{umap-software}
McInnes, L., Healy, J., Saul, N., \& Grossberger, L. (2018).  
UMAP: Uniform Manifold Approximation and Projection.  
\textit{The Journal of Open Source Software}, 3(29), 861.

\bibitem{UMAP}
McInnes, L., Healy, J., \& Melville, J. (2018, February).  
UMAP: Uniform Manifold Approximation and Projection for Dimension Reduction.  
\textit{arXiv preprint} arXiv:1802.03426.

\bibitem{kmeans}
MacQueen, J. (1967).  
Classification and analysis of multivariate observations.  
In \textit{5th Berkeley Symposium on Mathematical Statistics and Probability} (pp. 281–297).

\bibitem{Eigenbehaviour_water1}
Cominola, A., Spang, E. S., Giuliani, M., Castelletti, A., Lund, J. R., \& Loge, F. J. (2018).  
Segmentation analysis of residential water-electricity demand for customized demand-side management programs.  
\textit{Journal of Cleaner Production}, 172, 1607–1619.  
doi:10.1016/j.jclepro.2017.10.203.

\bibitem{Eigenbehavior_cognition}
Botros, A. A., Schuetz, N., Röcke, C., Weibel, R., Martin, M., Müri, R. M., \& Nef, T. (2022, April).  
Eigenbehaviour as an Indicator of Cognitive Abilities.  
\textit{Sensors}, 22(7), 2769. doi:10.3390/s22072769.

\bibitem{MDMS_news1}
Campbell, J. (2013, May 8).  
Meter Data Management System makes progress on installations, facilities.  
\textit{U.S. Army Engineering and Support Center Website}.  
Available: \url{https://www.hnc.usace.army.mil/Media/News-Stories/Article/482005/meter-data-management-system-makes-progress-on-installations-facilities/}.

\bibitem{Sec_Def_speech}
Vergun, D. (2021, April 22).  
Secretary of Defense Lloyd J. Austin III spoke today at the Leaders Summit on Climate.  
Available: \url{https://www.defense.gov/News/News-Stories/Article/Article/2582051/defense-secretary-calls-climate-change-an-existential-threat/}.

\bibitem{scikit-learn}
Pedregosa, F., Varoquaux, G., Gramfort, A., Michel, V., Thirion, B., Grisel, O., Blondel, M., Prettenhofer, P., Weiss, R., Dubourg, V., Vanderplas, J., Passos, A., Cournapeau, D., Brucher, M., Perrot, M., \& Duchesnay, E. (2011).  
Scikit-learn: Machine Learning in Python.  
\textit{Journal of Machine Learning Research}, 12, 2825–2830.

\bibitem{resilient_installations_article}
Surash, J. (2022, April 12).  
The push for resilient Army installations.  
\textit{U.S. Army Article}.  
Available: \url{https://www.army.mil/article/255603/the_push_for_resilient_army_installations}.

\bibitem{Army_installation_strat}
U.S. Army. (2020, December).  
Army Installations Strategy.  
\textit{U.S. Army Document}.



\end{thebibliography}
\end{document}